\documentclass[conference]{IEEEtran}
\IEEEoverridecommandlockouts

\usepackage{algorithm}
\usepackage{array}
\usepackage{textcomp}
\usepackage{stfloats}
\usepackage{url}
\usepackage{verbatim}
\usepackage{booktabs}
\usepackage{caption}
\usepackage{amsfonts}
\usepackage{hyperref}
\usepackage{color}
\usepackage{multirow}
\usepackage{makecell}
\usepackage{balance}

\usepackage{cite}
\usepackage{amsmath,amssymb,amsfonts}
\usepackage{algorithmic}
\usepackage{graphicx}
\usepackage{textcomp}
\usepackage{xcolor}
\def\BibTeX{{\rm B\kern-.05em{\sc i\kern-.025em b}\kern-.08em
    T\kern-.1667em\lower.7ex\hbox{E}\kern-.125emX}}
\begin{document}

\title{Soft Knowledge Distillation with Multi-Dimensional Cross-Net Attention for Image Restoration Models Compression\\
\thanks{This work is supported by National Key Research and Development Program of China (No. 2021YFB3101300).}
\thanks{\IEEEauthorrefmark{1} Corresponding author (D. Yan).}
}

\author{
\IEEEauthorblockN{
    Yongheng Zhang,
    and Danfeng Yan\IEEEauthorrefmark{1}}
    \IEEEauthorblockA{State Key Laboratory of Networking and Switching Technology, 
    BUPT, Beijing, China\\
    Email: {zhangyongheng, yandf}@bupt.edu.cn}
}

\maketitle

\begin{abstract}
Transformer-based encoder-decoder models have achieved remarkable success in image-to-image transfer tasks, particularly in image restoration. 
However, their high computational complexity—manifested in elevated FLOPs and parameter counts—limits their application in real-world scenarios.
Existing knowledge distillation methods in image restoration typically employ lightweight student models that directly mimic the intermediate features and reconstruction results of the teacher, overlooking the implicit attention relationships between them.
To address this, we propose a Soft Knowledge Distillation (SKD) strategy that incorporates a Multi-dimensional Cross-net Attention (MCA) mechanism for compressing image restoration models. 
This mechanism facilitates interaction between the student and teacher across both channel and spatial dimensions, enabling the student to implicitly learn the attention matrices. 
Additionally, we employ a Gaussian kernel function to measure the distance between student and teacher features in kernel space, ensuring stable and efficient feature learning.
To further enhance the quality of reconstructed images, we replace the commonly used L1 or KL divergence loss with a contrastive learning loss at the image level. 
Experiments on three tasks—image deraining, deblurring, and denoising—demonstrate that our SKD strategy significantly reduces computational complexity while maintaining strong image restoration capabilities.

\end{abstract}
\begin{IEEEkeywords}
Knowledge distillation, multi-dimensional cross-net attention, image restoration, contrastive learning
\end{IEEEkeywords}

\section{Introduction}
\label{sec:intro}

Image restoration models have significant deployment needs on edge devices such as self-driving cars, cellphones, and smart robots. 
However, the computational complexity and large parameter scales of existing models often exceed the capabilities of these mobile devices. 
This creates an urgent need for the compression of image restoration models, making it a critical area of research with important practical implications.

Model compression via knowledge distillation was first introduced by Hinton \textit{et al.} \cite{hinton2015distilling}, where student models primarily learn from the teacher’s logits. 
Since then, various distillation methods have been developed, focusing on responses \cite{meng2019conditional, zhang2019fast, muller2019does}, intermediate features \cite{romero2014fitnets, passalis2020heterogeneous, chen2021cross}, attention matrices \cite{zagoruyko2016paying, passban2021alp}, and instance relations \cite{chen2020learning, yu2019learning}. 
These approaches have been widely applied to detection and classification tasks.

In image-to-image transfer tasks, including image restoration, model compression methods based on knowledge distillation have only recently emerged.
Current approaches generally focus on learning from reconstructed images \cite{chen2020distilling, zhang2022wavelet} or a combination of intermediate features and reconstructed images \cite{gao2018image, fang2023dual, jin2021teachers}. 
However, these methods often overlook implicit attention relationships and may suffer from stability issues.

\begin{figure}[t]
\centering
    \includegraphics[width = 0.9\columnwidth]{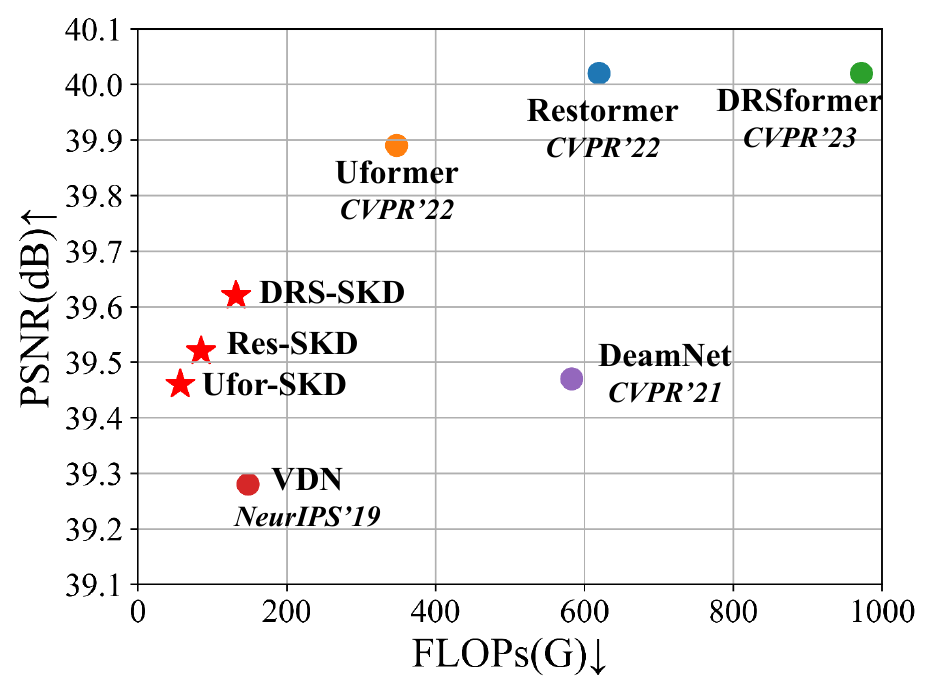}
    \caption{PSNR$\uparrow$ vs. FLOPs$\downarrow$ denoising results on SIDD.}
  \label{fig:first}
\end{figure}

To address these challenges, we propose a \textbf{S}oft \textbf{K}nowledge \textbf{D}istillation (\textbf{SKD}) strategy with \textbf{M}ulti-dimensional \textbf{C}ross-net \textbf{A}ttention (\textbf{MCA}) for compressing image restoration models. 
Our SKD strategy introduces key improvements: 
At the feature level, MCA enables interaction between student and teacher networks across channel and spatial dimensions, embedding attention relationships within the student features. 
Moreover, instead of directly mimicking teacher features, we employ Gaussian kernel functions to guide learning in kernel space, ensuring efficiency and stability. 
At the image level, we replace traditional L1 or KL divergence loss with contrastive learning loss, where the teacher's reconstructions serve as positive examples and degraded images as negatives, encouraging the student’s output to diverge from degraded instances.  
These innovations not only improve the student model's ability to learn complex relationships but also enhance its robustness across different degradation types.
Comparisons across multiple tasks and models confirm the superiority of our SKD strategy over other knowledge distillation-based compression methods and full restoration models.

\begin{figure*}[t]
\setlength{\abovecaptionskip}{0.3cm}
\setlength{\belowcaptionskip}{-0.3cm}
\centering
  \includegraphics[width=2.0\columnwidth]{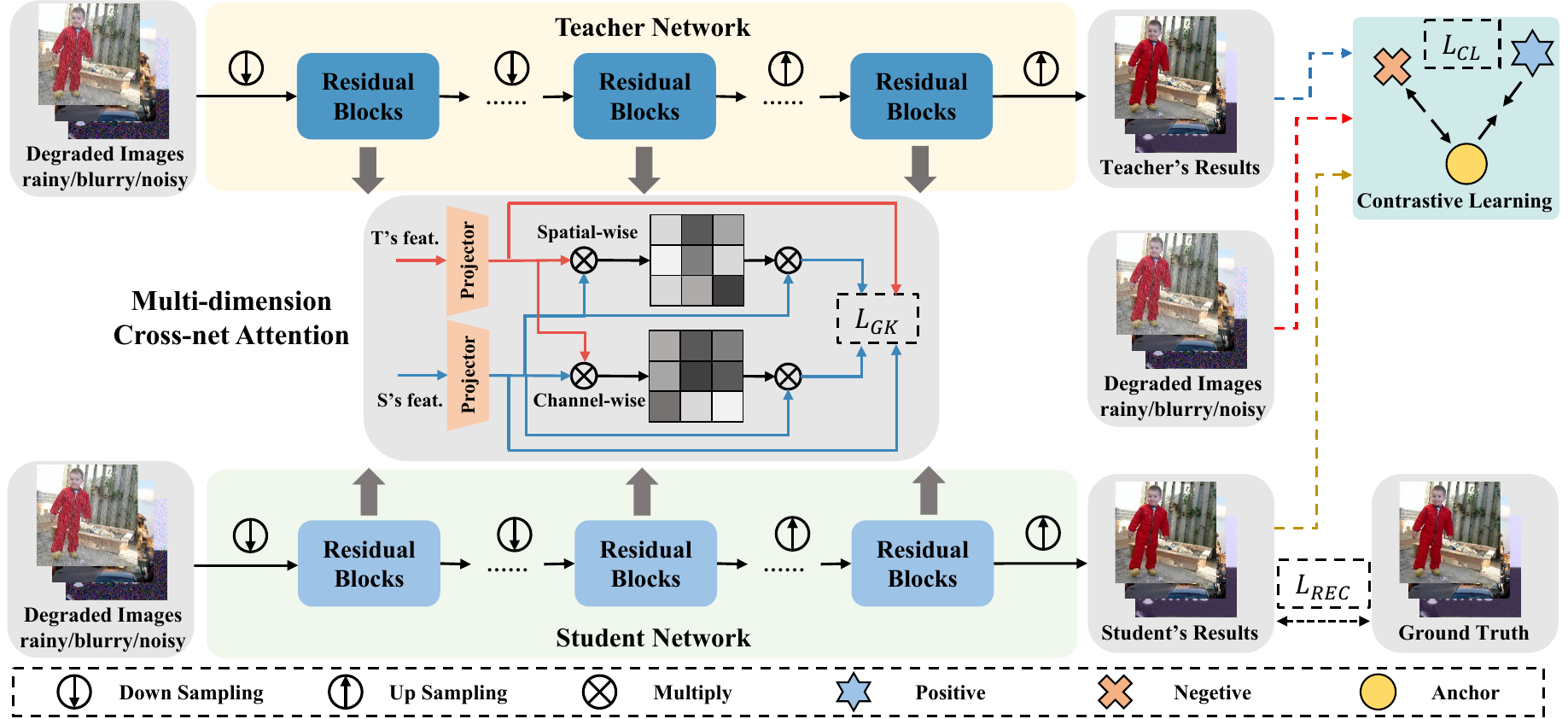}
  \caption{The overall architecture of the proposed Soft Knowledge Distillation (SKD) for image restoration models compression.}
  \label{fig:architecture}
\end{figure*}

\section{Proposed Method}
\label{sec:method}
\subsection{Overall Pipeline}
As illustrated in Fig. \ref{fig:architecture}, the SKD strategy utilizes a teacher-student network structure, where a degraded image $I\in\mathcal{R}^{H\times W \times3}$ is input into both networks. 
The pre-trained, complex teacher network excels at removing degradation factors, effectively restoring clean images. Meanwhile, the student network compresses model complexity by reducing the number of transformer layers and feature channels in its residual blocks. 
The student learns from the teacher at both the feature level and the reconstructed image level, ensuring it can achieve high performance despite its reduced size.

Feature-level learning is accomplished through our proposed Multi-dimensional Cross-network Attention (MCA) mechanism. 
The intermediate features of the student network interact with  features of corresponding blocks in the teacher network, allowing the student to implicitly absorb the attention knowledge embedded within the teacher.
The resulting student and teacher features are then mapped to Gaussian kernel space, and the loss is computed based on their distance, enabling stable and efficient knowledge transfer.

At the image level, in addition to the reconstruction loss computed with ground truth, contrastive learning helps the student further refine its output. 
The student’s reconstructed image uses the teacher’s output as a positive example, aligning closely with it, while multiple original degraded images serve as negative examples, encouraging divergence from these degraded instances.

These components of the SKD strategy work together to indirectly but significantly enhance the efficiency and stability of the student network during the distillation process, distinguishing our approach from direct imitation methods.

\subsection{Multi-dimension Cross-net Attention}
\label{ssec:MCA}
The proposed MCA mechanism facilitates interaction between student and teacher features across two dimensions: channel and spatial. 
Given the features from corresponding blocks in the teacher and student networks, we first use projectors to map these features into a unified dimensional space, represented as $T_f^i$ and $S_f^i$.
The interaction process, which yields the updated student features $S_{fc}^i$ (channel) and $S_{ft}^i$ (spatial), can be expressed as:
{\setlength\abovedisplayskip{5pt}
\setlength\belowdisplayskip{5pt}
\begin{equation}
\begin{aligned}
S_{fc}^i = softmax(T_f^i \cdot (S_f^i)^\mathsf{T} / \lambda) \cdot S_f^i,\\
S_{ft}^i = S_f^i \cdot softmax((T_f^i)^\mathsf{T} \cdot S_f^i / \lambda),
\end{aligned}
\end{equation}
}
where $\lambda$ is an optional temperature factor defined by $\lambda=\sqrt{d}$.
These features are subsequently mapped to Gaussian kernel space, where the Gaussian kernel distance and overall Gaussian kernel loss are calculated.
The Gaussian kernel distance and loss are defined as:
{\setlength\abovedisplayskip{5pt}
\setlength\belowdisplayskip{0pt}
\begin{equation}
\begin{aligned}
GK(x, y) = 1 - exp(-\frac{||x - y||_2^2}{2\sigma^2}),
\end{aligned}
\end{equation}
}
{\setlength\abovedisplayskip{0pt}
\setlength\belowdisplayskip{5pt}
\begin{equation}
\begin{aligned}
L_{GK} = GK(S_f^i, T_f^i)) + \alpha1(GK(S_{fc}^i, T_f^i) + GK(S_{ft}^i, T_f^i)),
\end{aligned}
\end{equation}
}
where $\sigma$ is the width of Gaussian kernel function. 
\subsection{Contrastive Learning for Knowledge Distillation}
\label{ssec:CL}
Contrastive learning, initially introduced for representation learning tasks, promotes an anchor point to move closer to positive example while distancing itself from negative ones \cite{gutmann2010noise, sohn2016improved}. 
Recently, this technique has been applied in various fields, including image restoration \cite{Yuntong2022UnsupervisedDW, wu2021contrastive}. 
We extend its application to knowledge distillation by using the student’s reconstructed images as anchors, the teacher’s outputs as positive examples, and a batch of degraded images as negative examples.
By minimizing the distance between the anchor and positive examples while maximizing the distance from negative examples, the student network learns to reconstruct clean images more effectively.
The contrastive learning loss $L_{CL}$ is formulated as:
{
\begin{equation}
\begin{aligned}
&L_{CL}(S_r, T_r, I) = \\
&-\log \frac{\operatorname{sim}\left(\phi(S_r), \phi(T_r)\right)}{\operatorname{sim}\left(\phi(S_r), \phi(T_r)\right)+\sum_{q=1}^{b} \operatorname{sim}\left(\phi(S_r), \phi(I^{q})\right)},
\end{aligned}
\end{equation}
}
where $S_r$, $T_r$, $I$ represent the student’s output, the teacher’s output (positive sample), and the degraded images (negative samples), respectively. 
The batch size is denoted by $b$, and $\operatorname{sim}(u, v)=\exp \left(\frac{u^{T} v}{\|u\|\|v\| \tau}\right)$ measures the similarity between two feature vectors, with $\tau$ as the temperature parameter and $\phi()$ representing a feature extraction operation using VGG-19 \cite{simonyan2014very}.

\begin{figure}[t]
\setlength{\abovecaptionskip}{0.1cm}
\setlength{\belowcaptionskip}{-0.5cm}
\centering
  \includegraphics[width=\columnwidth]{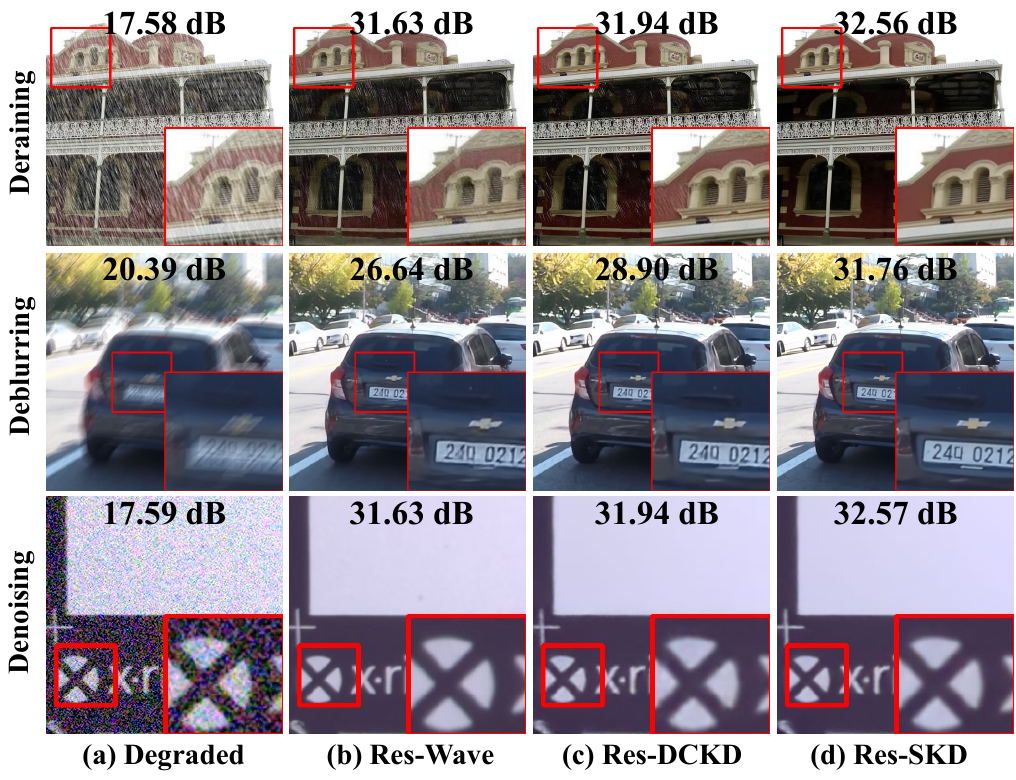}
  \caption{Qualitative results of knowledge distillation methods.}
  \label{fig:dis}
\end{figure}

\subsection{Overall loss}
\label{ssec:loss}
The reconstruction loss between student's results $S_r$ and ground truth $G$ is formulated as:
{\setlength\abovedisplayskip{5pt}
\setlength\belowdisplayskip{5pt}
\begin{equation}
\begin{aligned}
L_{REC} = ||G - S_r||_1.
\end{aligned}
\end{equation}
}
The overall loss is expressed as:
{\setlength\abovedisplayskip{5pt}
\setlength\belowdisplayskip{5pt}
\begin{equation}
\begin{aligned}
L = L_{REC} + \alpha2L_{GK} + \alpha3L_{CL},
\end{aligned}
\end{equation}
}
where $\alpha2$ and $\alpha3$ are trade-off weights.

\begin{figure}[t]
\setlength{\abovecaptionskip}{0.1cm}
\setlength{\belowcaptionskip}{-0.2cm}
\centering
  \includegraphics[width=\columnwidth]{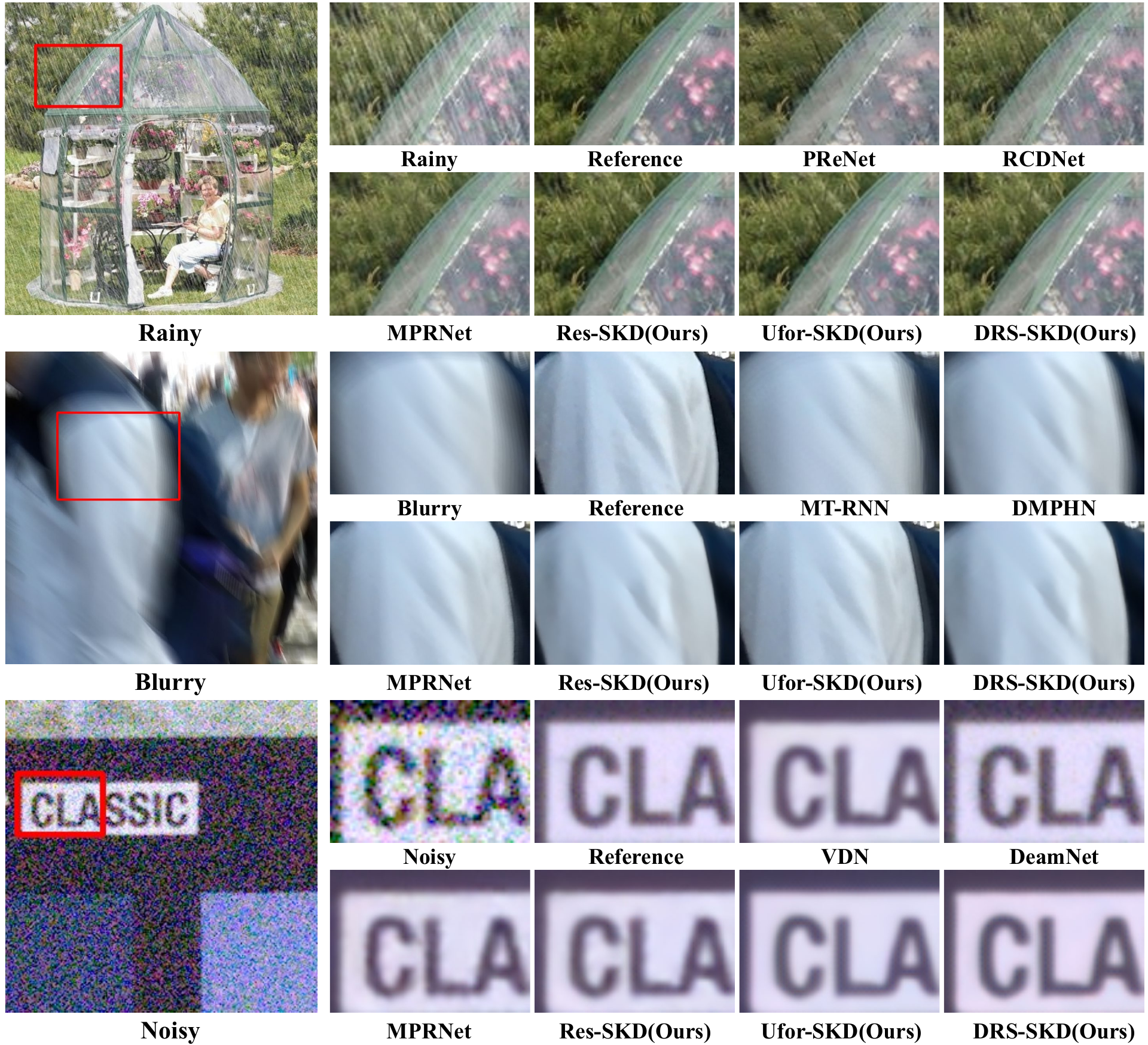}
  \caption{Qualitative comparison with light-weight methods.}
  \label{fig:main}
\end{figure}

\begin{table}[tp]
\small
\tabcolsep=4pt
\renewcommand\arraystretch{1.1}
\setlength{\abovecaptionskip}{0.1cm}
\setlength{\belowcaptionskip}{-0cm}
\centering
  \caption{Quantitative results of knowledge distillation methods across three task.}
  \label{tab:full-dis}
  \begin{tabular}{c|c|c|c|c|c|c}
    \hline
    Tasks &  \multicolumn{2}{c|}{Deraining} & \multicolumn{2}{c|}{Deblurring} & \multicolumn{2}{c}{Denoising} \\
    \hline
    Metrics & PSNR & SSIM & PSNR & SSIM & PSNR & SSIM\\
    \hline
    Restormer\cite{zamir2022restormer} & 33.69 & 0.935 & 32.07 & 0.952 & 40.02&0.960 \\
    Res-Wave \cite{zhang2022wavelet} & 32.27 & 0.918 & 30.36 & 0.933 & 39.25&0.951\\
    Res-DCKD \cite{fang2023dual} & 32.42 & 0.921 & 30.46 & 0.937 & 39.34&0.954\\
    Res-SKD &   \textbf{32.92} & \textbf{0.927} & \textbf{31.07} & \textbf{0.946} & \textbf{39.52}&\textbf{0.956}\\
    \hline
    Uformer\cite{wang2022uformer} & 33.34 & 0.931 & 31.98 & 0.960 & 39.89&0.960 \\
    Ufor-Wave \cite{zhang2022wavelet} & 32.10 & 0.917 & 30.50 & 0.938 & 39.26&0.951\\
    Ufor-DCKD \cite{fang2023dual} & 32.13 & 0.918 & 30.50 & 0.937 & 39.28&0.952\\
    Ufor-SKD &   \textbf{32.47} & \textbf{0.921} & \textbf{31.14} & \textbf{0.940} & \textbf{39.46}&\textbf{0.954}\\
    \hline
    DRSformer\cite{chen2023learning} & 33.82 & 0.937 & 31.97 & 0.949 & 40.03&0.960 \\
    DRS-Wave \cite{zhang2022wavelet} & 32.51 & 0.922 & 30.30 & 0.931 & 39.25&0.951\\
    DRS-DCKD \cite{fang2023dual} & 32.58 & 0.922 & 30.43 & 0.936 & 39.27&0.951\\
    DRS-SKD &  \textbf{33.11}  & \textbf{0.927} & \textbf{31.09} & \textbf{0.945} & \textbf{39.62}&\textbf{0.957} \\
    \hline
  \end{tabular}
\end{table}

\section{Experimental Results}
\label{sec:exp}
\subsection{Implementation Details} 
We evaluate our \textbf{S}oft \textbf{K}nowledge \textbf{D}istillation (\textbf{SKD}) using five datasets across three image restoration tasks:
Rain1400\cite{fu2017removing} and Test1200\cite{zhang2018density} for deraining, Gopro\cite{nah2017deep} and HIDE\cite{shen2019human} for deblurring, and SIDD\cite{abdelhamed2018high} for denoising.

For quantitative analysis of image quality, we employ two full-reference metrics: Peak Signal-to-Noise Ratio (PSNR) \cite{huynh2008scope} in dB, and Structural Similarity Index (SSIM) \cite{wang2004image}. 
To assess model complexity, we measure FLOPs and inference time on each $512 \times 512$ image. The best results are highlighted in bold, and the sub-optimal results are underlined.

The entire strategy is implemented in PyTorch, using Adam as the optimizer. 
The temperature parameter is set to $\tau = 1e-6$. 
The trade-off weights are $\alpha_{1} = 0.5$, $\alpha_{2} = 0.2$, and $\alpha_{3} = 0.2$. 
The student models are trained for 100 epochs with a batch size of 8. 
The learning rate starts at $2e-4$ and is gradually reduced to $1e-6$ using cosine annealing \cite{loshchilov2016sgdr}.
During training, all images are randomly cropped into $128 \times 128$ patches with pixel values normalized to [-1, 1].

For the teacher networks, we select three complex yet effective transformer-based models: Restormer \cite{zamir2022restormer}, Uformer \cite{wang2022uformer}, and DRSformer \cite{chen2023learning}. 
The number of layers in each level of the encoder-decoder and the dimensions of the teacher networks are \{\{4,6,6,8\}, \{1,2,8,8\}, \{4,4,6,6,8\}\} and \{48, 32, 48\}, respectively.  
The corresponding student models, Res-SKD, Ufor-SKD, and DRS-SKD, compress the hyper-parameters to \{\{1,2,2,4\}, \{1,2,4,4\}, \{2,2,2,2,4\}\} and \{32, 16, 32\}, resulting in 85.4\% and 85.8\% reduction of FLOPs and parameters, respectively.

\begin{table}[tp]
\small
\tabcolsep=1pt
\renewcommand\arraystretch{1.1}
\setlength{\abovecaptionskip}{0.1cm}
\setlength{\belowcaptionskip}{-0cm}
\centering
  \caption{Quantitative comparison with light-weight methods across three tasks.}
  \label{tab:light-dis}
  \begin{tabular}{c|c|c|c|c|c}
    \hline
Tasks&Deraining&Deblurring&Denoising&FLOPs & Infer time\\
    \hline
    PReNet\cite{ren2019progressive} & 31.56/0.914 &  -/- & -/- &176.7G&0.0589s\\
    RCDNet\cite{wang2020model} &   32.24/0.918 &  -/- & -/-&842.5G&0.1919s\\
    \hline
    DMPHN\cite{zhang2019deep}&  -/- & 30.14/0.932 &-/-&113.0G&0.0508s\\
    MT-RNN\cite{park2020multi} &  -/- & 30.15/0.931 &  -/- &579.0G&\underline{0.0387s}\\
    \hline
    VDN\cite{yue2019variational} &  -/- &  -/- & 39.28/0.956 &147.9G&0.0595s\\
    DeamNet\cite{ren2021adaptive} &   -/- & -/-&39.47/0.957&582.9G&0.0565s\\
    \hline
    MPRNet\cite{zamir2021multi} &  \textbf{33.28}/\textbf{0.927} &  \textbf{31.81}/\textbf{0.949} & \textbf{39.71}/\textbf{0.958}&565.0G&0.0593s\\
    \hline
    Res-SKD &   32.92/\textbf{0.927} &  31.08//\underline{0.946} &  39.52/0.956&\underline{85.0G}&\textbf{0.0356s}\\
    Ufor-SKD &   32.47/0.921 & \underline{31.14}/0.940 &  39.46/0.954&\textbf{57.1G}&0.0540s\\
    DRS-SKD &  \underline{33.11}/\textbf{0.927}  &  31.09/0.945 &  \underline{39.62}/\underline{0.957} &132.0G&0.0599s\\
    \hline
  \end{tabular}
\end{table}

\begin{figure}[t]
\setlength{\abovecaptionskip}{0.1cm}
\setlength{\belowcaptionskip}{-0.2cm}
\centering
    \includegraphics[width = \columnwidth]{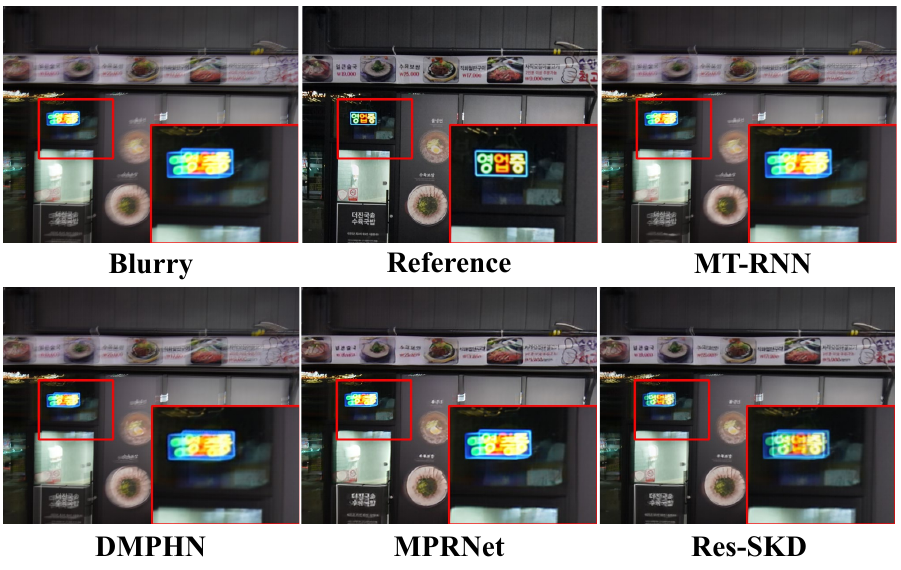}
    \caption{Deblurring results on real-world dataset BLUR-J\cite{rim2020real}.}
  \label{fig:real}
\end{figure}

\subsection{Comparisons with State-of-the-arts}
\textbf{Comparison with Knowledge Distillation methods.} We first compare our Soft Knowledge Distillation (SKD) strategy with two state-of-the-art (SOTA) image-to-image transfer knowledge distillation methods: Wave \cite{zhang2022wavelet} and DCKD \cite{fang2023dual}. 
Qualitative and quantitative results are presented in Fig. \ref{fig:dis} and Table \ref{tab:full-dis}.
The deraining and deblurring results in Table \ref{tab:full-dis} are averaged across the Rain1400\cite{fu2017removing} and Test1200\cite{zhang2018density}, Gopro\cite{nah2017deep} and HIDE\cite{shen2019human}, respectively.
Our distillation method significantly outperforms the other two SOTA methods in both visual quality of restored images and full-reference evaluation metrics.

\textbf{Comparison with Image Restoration methods.} We also compare our soft distillation strategy with seven image restoration methods, including two for deraining (PReNet \cite{ren2019progressive}, RCDNet \cite{wang2020model}), two for deblurring (DMPHN \cite{zhang2019deep}, MT-RNN \cite{park2020multi}), two for denoising (VDN \cite{yue2019variational}, DeamNet \cite{ren2021adaptive}), and one for generalized restoration (MPRNet \cite{zamir2021multi}). 
As shown in Fig. \ref{fig:main} and Table \ref{tab:light-dis},
our distilled models offer significantly lower complexity while achieving image quality and performance metrics comparable to complex models like MPRNet \cite{zamir2021multi}.

\textbf{Comparison on Real degraded Images.} We extended our evaluation to real blurry images, as shown in Fig. \ref{fig:real}.
Despite being trained on synthetic data, our distilled model Res-SKD effectively mitigates blur in real-world images.

\begin{table}[tp]
\small
\tabcolsep=7pt
\renewcommand\arraystretch{1.1}
\setlength{\abovecaptionskip}{0.1cm}
\setlength{\belowcaptionskip}{-0cm}
\centering
  \caption{Quantitative ablation study results on Gopro\cite{nah2017deep}.}
  \label{tab:abl}
  \begin{tabular}{c|c|c|c|c}
    \hline
    Sets & Channel-wise & Spatial-wise & $L_{CL}$ & PSNR/SSIM \\
    \hline
    (a) &  &  &  & 32.20/0.924 \\
    (b) & \checkmark &  &  & 32.61/0.929 \\
    (c) &  & \checkmark &  & 32.71/0.930 \\
    (d) & \checkmark & \checkmark &  & 32.99/0.933 \\
    \hline
    Res-SKD & \checkmark & \checkmark & \checkmark & \textbf{33.24/0.937} \\
    \hline
  \end{tabular}
\end{table}

\begin{figure}[t]
\setlength{\abovecaptionskip}{0.1cm}
\setlength{\belowcaptionskip}{-0.2cm}
\centering
    \includegraphics[width = \columnwidth]{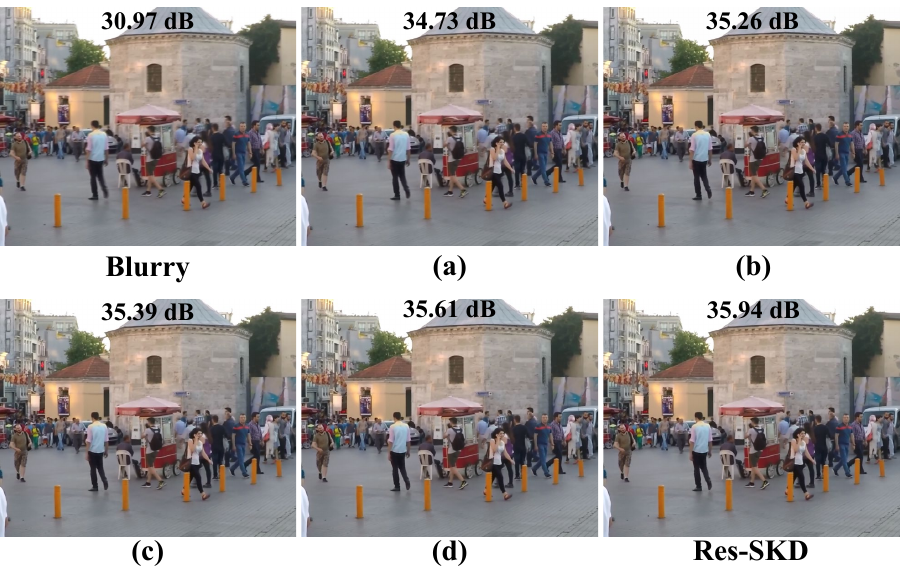}
    \caption{Qualitative ablation study results on Gopro\cite{nah2017deep}.}
  \label{fig:ablation}
\end{figure}

\subsection{Ablation Studies}
Ablation studies were conducted on the Gopro \cite{nah2017deep} dataset for deblurring 
with results summarized in Table \ref{tab:abl} and Fig. \ref{fig:ablation}. 
The channel-wise and spatial-wise attention mechanisms enhance the student model's ability to learn multi-dimensional knowledge from the teacher, leading to PSNR gains of 0.41 dB and 0.51 dB, respectively. 
The full Multi-dimensional Cross-net Attention (MCA) achieves a 0.79 dB increase in PSNR and a 0.009 improvement in SSIM over the baseline.
Additionally, the contrastive learning loss $L_{CL}$ contributes a further 0.25 dB gain in PSNR and a 0.004 improvement in SSIM. 
The qualitative results in Fig. \ref{fig:ablation} corroborate these findings, demonstrating the effectiveness of both the Multi-Dimensional Cross-Net Attention mechanism and contrastive learning loss in enhancing the distilled model's performance.

\section{Conclusion}
In this paper, we introduced a Soft Knowledge Distillation (SKD) strategy with a Multi-dimensional Cross-net Attention (MCA) mechanism to effectively compress transformer-based image restoration models. 
By enabling interaction between student and teacher networks across channel and spatial dimensions, our method allows the student model to implicitly learn attention matrices, ensuring efficient and stable feature learning. 
Additionally, we incorporated contrastive learning into the distillation process, with contrastive learning loss further improving the quality of reconstructed images. 
Experimental results on deraining, deblurring, and denoising tasks demonstrate that our SKD strategy significantly reduces computational complexity while maintaining high performance, making it ideal for real-world applications.

\newpage

\balance


\begin{thebibliography}{10}
\providecommand{\url}[1]{#1}
\csname url@samestyle\endcsname
\providecommand{\newblock}{\relax}
\providecommand{\bibinfo}[2]{#2}
\providecommand{\BIBentrySTDinterwordspacing}{\spaceskip=0pt\relax}
\providecommand{\BIBentryALTinterwordstretchfactor}{4}
\providecommand{\BIBentryALTinterwordspacing}{\spaceskip=\fontdimen2\font plus
\BIBentryALTinterwordstretchfactor\fontdimen3\font minus \fontdimen4\font\relax}
\providecommand{\BIBforeignlanguage}[2]{{%
\expandafter\ifx\csname l@#1\endcsname\relax
\typeout{** WARNING: IEEEtran.bst: No hyphenation pattern has been}%
\typeout{** loaded for the language `#1'. Using the pattern for}%
\typeout{** the default language instead.}%
\else
\language=\csname l@#1\endcsname
\fi
#2}}
\providecommand{\BIBdecl}{\relax}
\BIBdecl

\bibitem{zamir2021multi}
S.~W. Zamir, A.~Arora, S.~Khan, M.~Hayat, F.~S. Khan, M.-H. Yang, and L.~Shao, ``Multi-stage progressive image restoration,'' in \emph{Proceedings of the IEEE/CVF conference on CVPR}, 2021, pp. 14\,821--14\,831.

\bibitem{wang2022uformer}
Z.~Wang, X.~Cun, J.~Bao, W.~Zhou, J.~Liu, and H.~Li, ``Uformer: A general u-shaped transformer for image restoration,'' in \emph{Proceedings of the IEEE/CVF conference on CVPR}, 2022, pp. 17\,683--17\,693.

\bibitem{zamir2022restormer}
S.~W. Zamir, A.~Arora, S.~Khan, M.~Hayat, F.~S. Khan, and M.-H. Yang, ``Restormer: Efficient transformer for high-resolution image restoration,'' in \emph{Proceedings of the IEEE/CVF conference on computer vision and pattern recognition}, 2022, pp. 5728--5739.

\bibitem{chen2023learning}
X.~Chen, H.~Li, M.~Li, and J.~Pan, ``Learning a sparse transformer network for effective image deraining,'' in \emph{Proceedings of the IEEE/CVF Conference on Computer Vision and Pattern Recognition}, 2023, pp. 5896--5905.

\bibitem{ren2019progressive}
D.~Ren, W.~Zuo, Q.~Hu, P.~Zhu, and D.~Meng, ``Progressive image deraining networks: A better and simpler baseline,'' in \emph{Proceedings of the IEEE/CVF conference on computer vision and pattern recognition}, 2019, pp. 3937--3946.

\bibitem{wang2020model}
H.~Wang, Q.~Xie, Q.~Zhao, and D.~Meng, ``A model-driven deep neural network for single image rain removal,'' in \emph{Proceedings of the IEEE/CVF conference on computer vision and pattern recognition}, 2020, pp. 3103--3112.

\bibitem{park2020multi}
D.~Park, D.~U. Kang, J.~Kim, and S.~Y. Chun, ``Multi-temporal recurrent neural networks for progressive non-uniform single image deblurring with incremental temporal training,'' in \emph{European Conference on Computer Vision}.\hskip 1em plus 0.5em minus 0.4em\relax Springer, 2020, pp. 327--343.

\bibitem{zhang2019deep}
H.~Zhang, Y.~Dai, H.~Li, and P.~Koniusz, ``Deep stacked hierarchical multi-patch network for image deblurring,'' in \emph{Proceedings of the IEEE/CVF conference on computer vision and pattern recognition}, 2019, pp. 5978--5986.

\bibitem{yue2019variational}
Z.~Yue, H.~Yong, Q.~Zhao, D.~Meng, and L.~Zhang, ``Variational denoising network: Toward blind noise modeling and removal,'' \emph{Advances in neural information processing systems}, vol.~32, 2019.

\bibitem{ren2021adaptive}
C.~Ren, X.~He, C.~Wang, and Z.~Zhao, ``Adaptive consistency prior based deep network for image denoising,'' in \emph{Proceedings of the IEEE/CVF conference on computer vision and pattern recognition}, 2021, pp. 8596--8606.

\bibitem{fu2017removing}
X.~Fu, J.~Huang, D.~Zeng, Y.~Huang, X.~Ding, and J.~Paisley, ``Removing rain from single images via a deep detail network,'' in \emph{CVPR}.\hskip 1em plus 0.5em minus 0.4em\relax IEEE, 2017, pp. 3855--3863.

\bibitem{zhang2018density}
H.~Zhang and V.~M. Patel, ``Density-aware single image de-raining using a multi-stream dense network,'' in \emph{Proceedings of the IEEE conference on computer vision and pattern recognition}, 2018, pp. 695--704.

\bibitem{nah2017deep}
S.~Nah, T.~Hyun~Kim, and K.~Mu~Lee, ``Deep multi-scale convolutional neural network for dynamic scene deblurring,'' in \emph{Proceedings of the IEEE conference on computer vision and pattern recognition}, 2017, pp. 3883--3891.

\bibitem{shen2019human}
Z.~Shen, W.~Wang, X.~Lu, J.~Shen, H.~Ling, T.~Xu, and L.~Shao, ``Human-aware motion deblurring,'' in \emph{Proceedings of the IEEE/CVF international conference on computer vision}, 2019, pp. 5572--5581.

\bibitem{abdelhamed2018high}
A.~Abdelhamed, S.~Lin, and M.~S. Brown, ``A high-quality denoising dataset for smartphone cameras,'' in \emph{Proceedings of the IEEE conference on computer vision and pattern recognition}, 2018, pp. 1692--1700.

\bibitem{loshchilov2016sgdr}
I.~Loshchilov and F.~Hutter, ``Sgdr: Stochastic gradient descent with warm restarts,'' \emph{arXiv preprint arXiv:1608.03983}, 2016.

\bibitem{rim2020real}
J.~Rim, H.~Lee, J.~Won, and S.~Cho, ``Real-world blur dataset for learning and benchmarking deblurring algorithms,'' in \emph{ECCV}.\hskip 1em plus 0.5em minus 0.4em\relax Springer, 2020, pp. 184--201.

\bibitem{huynh2008scope}
Q.~Huynh-Thu and M.~Ghanbari, ``Scope of validity of psnr in image/video quality assessment,'' \emph{Electronics letters}, vol.~44, no.~13, pp. 800--801, 2008.

\bibitem{wang2004image}
Z.~Wang, A.~C. Bovik, H.~R. Sheikh, and E.~P. Simoncelli, ``Image quality assessment: from error visibility to structural similarity,'' \emph{IEEE transactions on image processing}, vol.~13, no.~4, pp. 600--612, 2004.

\bibitem{mittal2012no}
A.~Mittal, A.~K. Moorthy, and A.~C. Bovik, ``No-reference image quality assessment in the spatial domain,'' \emph{IEEE Transactions on image processing}, vol.~21, no.~12, pp. 4695--4708, 2012.

\bibitem{venkatanath2015blind}
N.~Venkatanath, D.~Praneeth, M.~C. Bh, S.~S. Channappayya, and S.~S. Medasani, ``Blind image quality evaluation using perception based features,'' in \emph{NCC}.\hskip 1em plus 0.5em minus 0.4em\relax IEEE, 2015, pp. 1--6.

\bibitem{hinton2015distilling}
G.~Hinton, O.~Vinyals, and J.~Dean, ``Distilling the knowledge in a neural network,'' \emph{arXiv preprint arXiv:1503.02531}, 2015.

\bibitem{romero2014fitnets}
A.~Romero, N.~Ballas, S.~E. Kahou, A.~Chassang, C.~Gatta, and Y.~Bengio, ``Fitnets: Hints for thin deep nets,'' \emph{arXiv preprint arXiv:1412.6550}, 2014.

\bibitem{zagoruyko2016paying}
S.~Zagoruyko and N.~Komodakis, ``Paying more attention to attention: Improving the performance of convolutional neural networks via attention transfer,'' \emph{arXiv preprint arXiv:1612.03928}, 2016.

\bibitem{kim2018paraphrasing}
J.~Kim, S.~Park, and N.~Kwak, ``Paraphrasing complex network: Network compression via factor transfer,'' \emph{Advances in neural information processing systems}, vol.~31, 2018.

\bibitem{li2020all}
R.~Li, R.~T. Tan, and L.-F. Cheong, ``All in one bad weather removal using architectural search,'' in \emph{Proceedings of the IEEE/CVF conference on CVPR}, 2020, pp. 3175--3185.

\bibitem{xie2023large}
C.~Xie, X.~Zhang, L.~Li, H.~Meng, T.~Zhang, T.~Li, and X.~Zhao, ``Large kernel distillation network for efficient single image super-resolution,'' in \emph{Proceedings of the IEEE/CVF Conference on Computer Vision and Pattern Recognition}, 2023, pp. 1283--1292.

\bibitem{zhang2022wavelet}
L.~Zhang, X.~Chen, X.~Tu, P.~Wan, N.~Xu, and K.~Ma, ``Wavelet knowledge distillation: Towards efficient image-to-image translation,'' in \emph{Proceedings of the IEEE/CVF conference on computer vision and pattern recognition}, 2022, pp. 12\,464--12\,474.

\bibitem{fang2023dual}
H.~Fang, Y.~Long, X.~Hu, Y.~Ou, Y.~Huang, and H.~Hu, ``Dual cross knowledge distillation for image super-resolution,'' \emph{Journal of Visual Communication and Image Representation}, vol.~95, p. 103858, 2023.

\bibitem{muller2019does}
R.~M{\"u}ller, S.~Kornblith, and G.~E. Hinton, ``When does label smoothing help?'' \emph{Advances in neural information processing systems}, vol.~32, 2019.

\bibitem{zhang2019fast}
F.~Zhang, X.~Zhu, and M.~Ye, ``Fast human pose estimation,'' in \emph{Proceedings of the IEEE/CVF conference on computer vision and pattern recognition}, 2019, pp. 3517--3526.

\bibitem{meng2019conditional}
Z.~Meng, J.~Li, Y.~Zhao, and Y.~Gong, ``Conditional teacher-student learning,'' in \emph{ICASSP 2019-2019 IEEE International Conference on Acoustics, Speech and Signal Processing (ICASSP)}.\hskip 1em plus 0.5em minus 0.4em\relax IEEE, 2019, pp. 6445--6449.

\bibitem{chen2021cross}
D.~Chen, J.-P. Mei, Y.~Zhang, C.~Wang, Z.~Wang, Y.~Feng, and C.~Chen, ``Cross-layer distillation with semantic calibration,'' in \emph{Proceedings of the AAAI conference on artificial intelligence}, vol.~35, no.~8, 2021, pp. 7028--7036.

\bibitem{passalis2020heterogeneous}
N.~Passalis, M.~Tzelepi, and A.~Tefas, ``Heterogeneous knowledge distillation using information flow modeling,'' in \emph{Proceedings of the IEEE/CVF conference on computer vision and pattern recognition}, 2020, pp. 2339--2348.

\bibitem{yu2019learning}
L.~Yu, V.~O. Yazici, X.~Liu, J.~v.~d. Weijer, Y.~Cheng, and A.~Ramisa, ``Learning metrics from teachers: Compact networks for image embedding,'' in \emph{Proceedings of the IEEE/CVF Conference on Computer Vision and Pattern Recognition}, 2019, pp. 2907--2916.

\bibitem{chen2020learning}
H.~Chen, Y.~Wang, C.~Xu, C.~Xu, and D.~Tao, ``Learning student networks via feature embedding,'' \emph{IEEE Transactions on Neural Networks and Learning Systems}, vol.~32, no.~1, pp. 25--35, 2020.

\bibitem{passban2021alp}
P.~Passban, Y.~Wu, M.~Rezagholizadeh, and Q.~Liu, ``Alp-kd: Attention-based layer projection for knowledge distillation,'' in \emph{Proceedings of the AAAI Conference on artificial intelligence}, vol.~35, no.~15, 2021, pp. 13\,657--13\,665.

\bibitem{jin2021teachers}
Q.~Jin, J.~Ren, O.~J. Woodford, J.~Wang, G.~Yuan, Y.~Wang, and S.~Tulyakov, ``Teachers do more than teach: Compressing image-to-image models,'' in \emph{Proceedings of the IEEE/CVF Conference on Computer Vision and Pattern Recognition}, 2021, pp. 13\,600--13\,611.

\bibitem{chen2020distilling}
H.~Chen, Y.~Wang, H.~Shu, C.~Wen, C.~Xu, B.~Shi, C.~Xu, and C.~Xu, ``Distilling portable generative adversarial networks for image translation,'' in \emph{Proceedings of the AAAI Conference on Artificial Intelligence}, vol.~34, no.~04, 2020, pp. 3585--3592.

\bibitem{gao2018image}
Q.~Gao, Y.~Zhao, G.~Li, and T.~Tong, ``Image super-resolution using knowledge distillation,'' in \emph{Asian Conference on Computer Vision}.\hskip 1em plus 0.5em minus 0.4em\relax Springer, 2018, pp. 527--541.

\bibitem{simonyan2014very}
K.~Simonyan and A.~Zisserman, ``Very deep convolutional networks for large-scale image recognition,'' \emph{arXiv preprint arXiv:1409.1556}, 2014.

\bibitem{gutmann2010noise}
M.~Gutmann and A.~Hyv{\"a}rinen, ``Noise-contrastive estimation: A new estimation principle for unnormalized statistical models,'' in \emph{Proceedings of the thirteenth international conference on artificial intelligence and statistics}.\hskip 1em plus 0.5em minus 0.4em\relax JMLR Workshop and Conference Proceedings, 2010, pp. 297--304.

\bibitem{sohn2016improved}
K.~Sohn, ``Improved deep metric learning with multi-class n-pair loss objective,'' \emph{Advances in neural information processing systems}, vol.~29, 2016.

\bibitem{wu2021contrastive}
H.~Wu, Y.~Qu, S.~Lin, J.~Zhou, R.~Qiao, Z.~Zhang, Y.~Xie, and L.~Ma, ``Contrastive learning for compact single image dehazing,'' in \emph{Proceedings of the IEEE/CVF Conference on Computer Vision and Pattern Recognition}, 2021, pp. 10\,551--10\,560.

\bibitem{Yuntong2022UnsupervisedDW}
Y.~Ye, C.~Yu, Y.~Chang, L.~Zhu, X.-L. Zhao, L.~Yan, and Y.~Tian, ``Unsupervised deraining: Where contrastive learning meets self-similarity,'' in \emph{Proceedings of the IEEE/CVF Conference on Computer Vision and Pattern Recognition}, 2022, pp. 5821--5830.

\end{thebibliography}

\end{document}